\documentclass{article}

\usepackage{PRIMEarxiv}

\usepackage[utf8]{inputenc}
\usepackage[T1]{fontenc}
\usepackage{hyperref}
\usepackage{url}
\usepackage{booktabs}
\usepackage{amsfonts}
\usepackage{nicefrac}
\usepackage{microtype}
\usepackage{xcolor}
\usepackage{fancyhdr}
\usepackage{graphicx}
\usepackage{tikz}
\usetikzlibrary{positioning}
\usepackage[numbers,compress]{natbib}
\graphicspath{{media/}}

\title{Interpreting at Write Time: A Policy Ablation for Multi-Goal Agent Memory}

\author{%
  Albert Sadowski \\
  Faculty of Electronics and Information Technology \\
  Warsaw University of Technology \\
  Warsaw, Poland \\
  \texttt{albert.sadowski.stud@pw.edu.pl} \\
  \And
  Jaros{\l}aw A. Chudziak \\
  Faculty of Electronics and Information Technology \\
  Warsaw University of Technology \\
  Warsaw, Poland \\
  \texttt{jaroslaw.chudziak@pw.edu.pl} \\
}

\begin{document}

\maketitle

\begin{abstract}
A long-running assistant cannot keep everything it has seen, so it summarises. Summarising is not
neutral: what is kept depends on what the record is for, and the choice is made once, before anyone
knows which of the user's goals will ask. We ask what a memory should summarise for when it serves
several standing goals at once. Three write policies answer differently: summarise with no goal in
view, write one summary covering every goal, or write one summary per goal and read them together. We
compare them with five models, two short event streams and one LLM judge, holding the read step
fixed. Answers from different goal stores overlap less than answers from one store on two runs, so
the goals do pull apart. Per-goal summaries read together scored highest. They beat the
all-goal summary on relevance, completeness and accuracy, but the neutral one only on relevance and
completeness, and at matched answer length only on relevance. The all-goal summary scored below the
neutral one, which had a quarter of its budget. A single goal store served its own goal's queries
better than the other goal stores, but not better than the neutral summary.
\end{abstract}


\section{Introduction}
\label{sec:intro}

A long-running assistant cannot keep everything it has seen. Context is finite, and retrieval over a
growing raw history runs into the familiar limits of information retrieval, so systems compress, most
often by summarising. MemGPT summarises when the context
window fills up \citep{packer2023MemGPT}, MemoryBank writes daily summaries and keeps a running
portrait of the user \citep{zhong2023MemoryBank}, Generative Agents write reflections over an
observation stream
\citep{park2023GenerativeAgents}, and A-Mem writes structured notes and revises them as new events
arrive \citep{xu2025AMem}. Zep extracts entities and facts into a graph \citep{rasmussen2025Zep},
which commits the same way, since its schema decides what can be represented and must be fixed
before anyone knows which use will matter \citep{jiang2025Evolution}.

Summarising is not neutral. To compress you have to decide what matters, and that decision is made
against some notion of what the record is for. A summary is therefore written from a perspective,
whether or not the system names one. Call that choice the encoding.

The same stored summary then gets asked very different things. LongMemEval defines five separate
long-term memory abilities over one chat history \citep{wu2025LongMemEval}, LoCoMo asks for
question answering, event summarisation and dialogue generation over the same conversations
\citep{maharana2024LoCoMo}, and MemoryCD tests whether one person's preferences hold while their
interactions span twelve domains at once \citep{zhang2026MemoryCD}. One write, many uses. The
field's answer has been to split the storage: by memory type
\citep{sumers2024CognitiveArchitectures}, by episodic against semantic content
\citep{zhang2026MultiMemory}, and by task against environment to cut retrieval noise
\citep{gao2025EfficientContext}. All of that decides where a fact is filed. None of it decides what
the summary is written for.

Not all of that variety strains the encoding. A question about timing and one that chains two facts
use the same record differently but agree on what belongs in it, so one accurate summary serves
both. Goals are the hard case. A standing goal is a
lasting concern the assistant serves across sessions, not a one-off request: a person has their
money, their health, their relationships, their career. Goals rarely disagree about what happened.
They disagree about which parts of it were worth the space. A 15 percent discount granted during a
negotiation is a margin reduction, a pricing precedent, a trust signal and evidence of competitive
pressure, and a summary that renders it as one may render it poorly as the others. The encoding
decides this once, before any goal has asked. It matters when the history is longer than what can be
re-read: the summary then replaces the stream, and what it left out is gone however good the
retriever is.

Two questions follow, both about the write rather than the read. First, if a summary is written
for one goal, how does that affect what can be recovered from it later? The alternative is to
summarise with no goal in view and work out what the events mean at read time. Second, when the
assistant serves several goals at once, should it keep one summary per goal or one summary
covering all of them? Writing per goal is not filing per goal. Each goal's summary covers the
whole event, every summary is read at query time, and nothing is routed or split; the stores
differ only in the instruction that wrote them, which is the goal-conditioning. Rashomon Memory,
our earlier work \citep{rashomonMemory}, proposed the per-goal policy and showed it was feasible on
a single scenario, without comparing it against the alternatives.

This paper contributes that comparison, not the policy. We set four goals against one client
engagement, an account-management setting where the goals are explicit and the events are dense.
The comparison only means something if the four goals really do pull apart, so we measure that first
rather than assume it. Standing goals are a construct we define rather than one a benchmark
supplies; MemoryCD's twelve domains are topics, not goals. The design draws on encoding specificity
\citep{tulving1973encoding} and transfer-appropriate processing \citep{morris1977levels}, by which a
trace formed for one purpose serves that purpose best, and on Minsky's mind of competing specialists
\citep{minsky1986SocietyOfMind}. They motivate the hypothesis; the experiment does not test them.

\section{Write policies and setup}
\label{sec:setup}

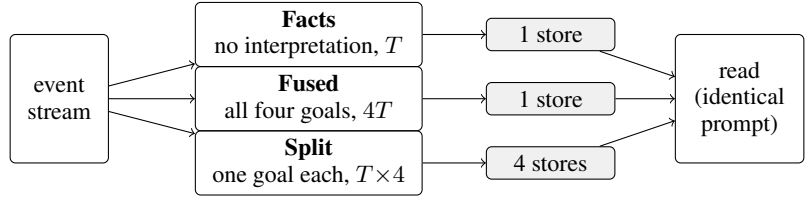
\begin{figure}[t]
\centering
\begin{tikzpicture}[font=\small,
ev/.style={draw,rounded corners=2pt,align=center,inner sep=3pt}, w/.style={draw,rounded
corners=2pt,align=center,inner sep=3pt,minimum width=30mm}, s/.style={draw,rounded
corners=2pt,fill=black!6,align=center,inner sep=3pt,minimum width=17mm}]
  \node[ev,minimum height=17mm,minimum width=13mm] (ev) at (0,0) {event\\stream};
  \node[w] (wa) at (3.3,0.85) {\textbf{Facts}\\no interpretation, $T$};
  \node[w] (wb) at (3.3,0.00) {\textbf{Fused}\\all four goals, $4T$};
  \node[w] (wc) at (3.3,-0.85) {\textbf{Split}\\one goal each, $T{\times}4$};
  \node[s] (sa) at (6.5,0.85) {1 store};
  \node[s] (sb) at (6.5,0.00) {1 store};
  \node[s] (sc) at (6.5,-0.85) {4 stores};
  \node[ev,minimum height=17mm,minimum width=17mm] (rd) at (9.0,0) {read\\(identical\\prompt)};
  \foreach \r in {a,b,c}{
    \draw[->] (ev) -- (w\r); \draw[->] (w\r) -- (s\r); \draw[->] (s\r) -- (rd);
}
\end{tikzpicture}
\caption{The three write policies. Only the write differs. The reader is the same model with the
same prompt throughout, receives the four stores concatenated under Split, and is never told which
goal a query is aligned with, so it cannot route to a store.}
\label{fig:policies}
\end{figure}

Let $T$ be the write budget (Figure~\ref{fig:policies}). Facts uses one encoder at budget $T$,
instructed to summarise what happened, who was involved and which numbers were stated, without
inferring meaning or assessing implications. Fused uses one encoder at budget $4T$, instructed to
cover in one pass what each of the four goals would find relevant. Split uses four encoders, one
per goal, each at budget $T$. Split and Fused thus share a nominal $4T$, and Facts has a quarter of
it. Budgets are prompt instructions, not enforced limits, and only answers were saved, so we cannot
confirm that store lengths matched. Split also makes four writes per event where Fused makes one.
Each goal encoder, like Facts, is asked for two to four sentences per event, Fused for four to six.
Both differences may favour Split.

The four goals are Financial, Risk, Relationship and Competitive. Each event is encoded under the
policy's instruction, then merged into the existing store.

We use two multi-session event streams from a long-running client engagement. Meridian has 5
events and 4 queries; Helix has 15 events and 6 queries, with mixed signals from several
stakeholders. Both are structured records rather than chat transcripts, which isolates the write
policy from conversational dynamics such as grounding and repair.

Five models act as encoder and reader (Claude Haiku 4.5, Claude Sonnet 4.5, GPT-5 mini, GPT-5
nano, Minimax M2.1), at temperature 0 where supported, running both streams under every policy ten
times. A judge (GPT-5 mini), blind to policy, scores every answer against the raw events on
relevance, completeness, accuracy and clarity, each from 1 to 5, which correlates with human
judgment in prior work \citep{zheng2023judging, liu2023geval}. The composite score is their mean.
The unit of analysis is a cell, one (model, stream, query) triple averaged over its ten runs,
giving 50 cells; run-to-run standard deviation inside a cell averages 0.60, which is why we
average. Every policy is observed in every cell, so we use paired tests throughout. Two leakage
controls matter: the Fused instruction names all four goals, so its store carries the goal
vocabulary the four goal stores carry between them, and the reader never sees a goal label. Code
and all scored answers are at \url{https://github.com/albsadowski/goal-conditioned-encoding}.

To ask whether the four goal records diverge at all, we compare answers by content-word Jaccard
overlap, lower-casing and dropping stopwords. An answer is what a reader gets from a store, so
overlap between two answers stands in for overlap between two records. The noise floor is the
overlap between two answers from the \emph{same} goal store on two independent runs.

\section{Results}
\label{sec:results}

We start with the premise. Answers drawn from two different goal stores in the same run overlap at
$0.121$; answers drawn from the same goal store on two independent runs overlap at $0.190$. The across-goal figure is lower in all 50 cells,
$d_z = 1.95$, $t(49) = 13.80$, $p < .001$, and lower in every model and in both streams. The
comparison is conservative: the floor absorbs a full re-encode and a fresh answer while the
across-goal figure holds the run fixed, and the four stores answer at similar lengths. At least in
wording, writing one event for four goals separates it further than writing it twice for a single
goal does.

\begin{table}[t]
\centering
\caption{Mean scores by write policy over 50 cells (model $\times$ stream $\times$ query, each
averaged over 10 runs). Scores run from 1 to 5. Budgets are instructed, not measured. Best-single
is a post-hoc oracle: the best of the four goal stores, each written at budget $T$, chosen once the
scores are known.}
\label{tab:main}
\small
\begin{tabular}{@{}lccccccc@{}}
\toprule
\textbf{Policy} & \textbf{Budget} & \textit{M} & \textit{SD} & Rel. & Compl. & Acc. & Clar. \\
\midrule
Split (four goal writes)     & $4T$ & 3.97 & 0.77 & 4.42 & 3.45 & 3.50 & 4.52 \\
Best-single (oracle)         & $T$  & 3.79 & 0.88 & --   & --   & --   & --   \\
Facts (no interpretation)    & $T$  & 3.68 & 1.06 & 3.73 & 2.97 & 3.46 & 4.54 \\
Fused (all goals, one write) & $4T$ & 3.34 & 0.85 & 3.61 & 2.65 & 2.74 & 4.34 \\
\bottomrule
\end{tabular}
\end{table}

Table~\ref{tab:main} reports the policies. The cleanest test of what a summary should be written
for is Facts against Fused, since both are single encoders making one pass into one store. Facts ($M = 3.68$) scored above Fused ($M = 3.34$): $+0.34$,
$d_z = 0.41$, $t(49) = 2.92$, $p = .005$, better in 34 of 50 cells, and it did so on a quarter of
the nominal budget. Accuracy accounts for most of the gap ($+0.72$, $t(49) = 4.63$, $p < .001$),
completeness for some of it ($+0.33$, $p = .021$), and relevance for none ($+0.12$, $p = .457$). So
answers from the all-goal store were not only less complete. They were judged less faithful to the
events.

If the instruction matters, a store written under one goal should answer that goal's queries better
than stores differing from it in nothing else. Eight of the ten queries
carry a descriptive goal label in the scenario file, entering no prompt, the two labelled
ambiguous being excluded. For each we compare the store written under the matching goal against
the mean of the other three, identical in budget, architecture and read prompt. The aligned store
scored $3.68$ against $3.15$: $+0.53$, $d_z = 0.65$, $t(39) = 4.11$, $p < .001$, better in 31 of
40 cells, holding in all five models ($+0.36$ to $+0.74$) and both streams. Facts scored $3.93$ on
the same cells, so the aligned store beat the other goal stores but not the neutral summary
($-0.25$, $t(39) = -2.11$, $p = .042$).

That advantage is not spread evenly. It is largest for Competitive ($+1.25$) and Relationship
($+0.68$), and close to nothing for Risk ($+0.11$) and Financial ($+0.08$). Those are the same
goals ranked in reverse by how well a neutral summary already serves them: on their own labelled
queries Facts scored $3.45$ on Competitive and $3.89$ on Relationship, against $3.96$ on Financial
and $4.44$ on Risk, and only on Competitive did the aligned store beat Facts. One reading is that
where a query asks for details stated outright, the neutral summary has already kept them, and where
it asks for a reading of what happened, writing for the goal helps most. With two labelled queries
per goal, this is a pattern to test, not a finding.

Splitting the write across the four goals produced the largest policy effect in the study. Split
($M = 3.97$) beat Fused by $+0.64$, $d_z = 1.31$, $t(49) = 9.28$, $p < .001$, better in 46 of 50
cells at the same nominal budget, with the gain appearing on relevance ($+0.82$), completeness
($+0.80$) and accuracy ($+0.75$) alike. But Split makes four writes per event and is asked for more
text, so this contrast does not isolate the goal instruction. Against Facts the composite gap is
$+0.30$, $d_z = 0.40$, $t(49) = 2.82$, $p = .007$, but it is made of different material: relevance
($+0.70$, $p < .001$) and completeness ($+0.48$, $p < .001$) carry all of it, while accuracy
($+0.03$, $p = .823$) and clarity ($-0.03$, $p = .679$) are level. Split also had four times the
nominal budget of Facts.

The gain comes from reading across the stores, not from owning a good one. Split also beat
Best-single, the oracle picking whichever goal store scored highest once the scores are known
($+0.18$, $d_z = 0.41$, $p = .005$), and that oracle was itself only level with Facts ($+0.11$, $p
= .292$). The four goal stores alone scored $2.87$ to $3.51$, every one below Facts. A single
goal-conditioned summary was a worse memory than the neutral one here; only the four read together
scored above it.

Four checks bear on these contrasts. GPT-5 mini is also the judge, and dropping it from the encoder
set leaves Split over Fused at $+0.67$ ($t(39) = 8.08$, $p < .001$), Split over Facts at $+0.40$
($p = .003$) and Facts over Fused at $+0.27$ ($t(39) = 1.94$, $p = .060$). This tests
self-preference, not the judge's validity. Split and Facts each beat Fused in all five models, and
Split beat Facts in four, all but GPT-5 mini. LLM judges can favour longer answers, and Split writes
the longest (192 words, against 147 for Fused and 119 for Facts), so we regressed each within-cell
score difference on the word-count difference and read the intercept, the difference expected at
zero length gap. Split over Fused is unchanged ($+0.55$, $t(48) = 6.16$, $p < .001$) and Facts over
Fused rises to $+0.52$ ($t(48) = 4.30$, $p < .001$). For Split over Facts the gap at matched length
falls to $+0.11$ ($p = .391$), and only relevance still carries a difference ($+0.49$, $p = .012$).
Finally, with the query as the unit and differences averaged over models, Split over Fused holds in
all ten queries ($t(9) = 10.26$, $p < .001$), Split over Facts in eight ($p = .035$, before length
adjustment) and the aligned-store advantage in seven of eight ($t(7) = 2.43$, $p = .045$), while
Facts over Fused ($p = .055$) and Split over Best-single ($p = .059$) fall short of $.05$.

\section{Conclusion}
\label{sec:conclusion}

We compared three ways to write a memory that serves several standing goals. The goal stores gave
measurably different answers, which is what makes the choice a real one. Reading the per-goal
summaries together had the highest mean score. Its lead over the all-goal summary held across
models, streams and queries. Its lead over the neutral summary was smaller, came from relevance and
completeness, and was not significant at matched answer length. The all-goal summary was the
weakest in every model, behind even the one written with no goal in view, mostly on judged accuracy.

On average, no single goal store matched the neutral summary, even on its own goal's queries,
though it beat the other goal stores on them. That leaves the instruction given to the writer as a
design parameter in its own right, separate from how memory is partitioned or retrieved. Whether
writing per goal beats writing neutrally, once the amount written is held fixed, is still open.

\paragraph{Limitations.} Budgets were instructions, and Facts was not run
at $4T$, so we cannot separate the write policy from how much text each policy kept. Split also
differs from Fused in the number of writes, not only in what each write is for, and decomposing a
task into focused subtasks can help for reasons unrelated to goals \citep{kostka2025towards}.
The data are small: ten queries from two short streams, which we wrote so that the goals would
diverge. The 50 cells are therefore not 50 independent scenarios. One LLM judge scored answer
quality rather than downstream utility, with no human validation. The streams are
structured records rather than live dialogue, and the goal set was fixed in advance, so a goal with
no store is simply not retained.

\end{document}